\PassOptionsToPackage{hyphens,spaces,obeyspaces}{url}
\PassOptionsToPackage{dvipsnames}{xcolor}
\documentclass{article}

\usepackage{geometry}
\usepackage[utf8]{inputenc}
\usepackage[T1]{fontenc}
\usepackage{url}
\usepackage{booktabs}
\usepackage{amsfonts} 
\usepackage{amsmath}
\usepackage{amssymb}
\usepackage[prologue]{xcolor}
\usepackage[bookmarksnumbered,unicode]{hyperref}
\hypersetup{
  colorlinks=true,
  linkcolor=Blue,    
  citecolor=Blue,       
  urlcolor=Blue,     
  filecolor=Blue,    
  anchorcolor=Blue   
}
\hypersetup{
  pdflang={en},
  pdfdisplaydoctitle
}

\usepackage{graphicx}
\usepackage{nicefrac}
\usepackage{tabularx}
\usepackage{fancyhdr}
\RequirePackage[
  datamodel=acmdatamodel,
  style=acmauthoryear,
  backend=biber,
  giveninits=true,
  uniquename=init,
  mincrossrefs=99
]{biblatex}

\usepackage[american]{babel} 

\addto\extrasamerican{
}

\title{Zarya: A Hybrid Autoregressive--Masked Diffusion Language Model with Flexible Training and Dual-Mode Inference}

\author{%
Leonid Sinev,
Ilya Koziev,
and
Vladislav Leshchuk%
}

\date{}

\def\github{\raisebox{-1.5pt}{\includegraphics[height=1.05em]{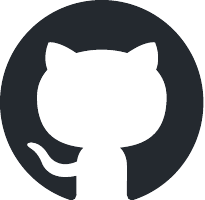}}}

\newcommand{\hflink}{https://huggingface.co/collections/ai-forever/zarya}
\newcommand{\ghlink}{https://github.com/ai-forever/zarya}

\begin{document}
\fancyhead[L,C]{}
\fancyhead[R]{Preprint. Please cite the peer-reviewed version when it is published.}
\fancyfoot[L]{}
\fancyfoot[C]{\thepage}
\fancyfoot[R]{}

\maketitle
\thispagestyle{fancy}

\begin{center}
\vspace{-0.5cm}
\begin{tabular}{rl}
\github & \url{\ghlink}\\
\end{tabular}
\end{center}

\begin{abstract}
Autoregressive language models (ARMs) are constrained by sequential,
left-to-right generation, while masked diffusion models (MDMs) enable
parallel decoding but suffer from high computational overhead due to the inability to reuse Key-Value (KV) cache and from incoherent
generation arising from learning dependencies over an intractable
space of token combinations.
We introduce \textsc{Zarya}, a family of
hybrid language models that jointly optimizes an autoregressive (AR)
objective and a masked-diffusion objective within a single
architecture.
\textsc{Zarya} structures training data into variable-size
\emph{slots} and employs a curriculum that gradually increases slot
granularity, enabling a smooth transition from fine-grained AR
learning to coarse-grained diffusion learning.
At inference,
\textsc{Zarya} provides two distinct decoding paradigms through a
unified interface: (i) MDM sampling with first-hitting
denoising, and (ii) \emph{slotted speculative decoding} that
interleaves inter-slot diffusion-based selection with intra-slot
autoregressive infilling, achieving full KV cache reuse.
The training and
inference regimes are fully decoupled, allowing a model trained with
any configuration to be deployed in either mode.
Extensive
configurability --- including grouped noise patterns (Prefix
Completion, Fill-In-the-Prefix, Fill-In-the-Middle), ordered sampling
schedules, and noise-level permutation strategies --- enables flexible
research exploration.
We release Zarya models publicly in sizes 0.6B, 1.7B, and 4B, demonstrating performance on standard benchmarks while offering a principled integration of autoregressive and diffusion paradigms.
\end{abstract}

\section{Introduction}

Autoregressive models (ARMs) have achieved remarkable success in
a wide range of natural language tasks~\cite{brown2020language,
openai2023gpt4, touvron2023llama2}.
However, their sequential,
left-to-right decoding fundamentally limits inference throughput,
preventing parallelization~\cite{chen2023speculative,
cai2024medusa}.
Masked diffusion models (MDMs) offer a compelling
alternative by enabling parallel generation through an iterative
denoising process without a fixed generation order~\cite{li-etal-2026-SurveyDiffusionLanguage}.
However, most MDMs suffer from two critical drawbacks: (i) they
exclude Key-Value (KV) caching, incurring high computational
overhead during inference; and (ii) they learn dependencies over an
intractable space of token combinations, leading to incoherent
generation~\cite{li2026refusion}.

Recent hybrid approaches have sought to bridge these paradigms.
Block Diffusion (\textsc{BD3-LMs})~\cite{arriola-etal-2025-BlockDiffusionInterpolating} interpolates
between AR and MDM by grouping tokens into fixed-size blocks and
unmasking them from left to right, but the block size is fixed and lacks
flexibility.
Esoteric Language Models (\textsc{Eso-LMs})~\cite{sahoo2026esoteric}
fuse AR and MDM paradigms at the loss level, enabling KV caching for
MDMs and achieving faster inference than contemporary MDMs.
\textsc{ReFusion}~\cite{li2026refusion} elevates parallel decoding
from the token level to a higher \emph{slot level}, interleaving
inter-slot diffusion-based selection with intra-slot autoregressive
infilling, outperforming Qwen3-8B~\cite{qwen3technicalreport}
on GSM8K~\cite{cobbe2021trainingverifierssolvemath}
and MBPP~\cite{austin-etal-2021-ProgramSynthesisLarge}
while being 2.33{\texttimes} faster on average.

We introduce \textsc{Zarya}, a novel hybrid architecture with the following key contributions:
\begin{enumerate}
\item \textbf{Training with gradually increasing slot length:}
  We partition the
  predicted sequence into fixed-length, consecutive sub-sequences, referred to as slots.
  The size of the slots gradually increased during training.
  For each sequence,
  we randomly mask several slots, reorder
  the input so that clean slots precede
  masked ones, and also permute the original order of both masked and clean slots.
  The
  model simultaneously learns AR next-token prediction on visible
  slots for sequential generation, and a
  denoising loss on the masked slots for
  context-aware parallel reconstruction.
\item \textbf{Training--inference decoupling:} Unlike prior work
  where the training configuration dictates the inference mode,
  \textsc{Zarya} allows any trained model to be deployed in either
  MDM sampling or slotted speculative decoding mode via a single
  inference flag, offering higher flexibility.
  Both modes are fully using KV cache with causal attention masks
\end{enumerate}

\section{Related Work}

\paragraph{Block Diffusion (BD3-LMs).}
\textsc{BD3-LMs} interpolate between AR and MDM by grouping tokens into
blocks and unmasking them left-to-right~\cite{arriola-etal-2025-BlockDiffusionInterpolating}.
However, the block size is fixed, limiting flexibility.

\paragraph{Esoteric Language Models (Eso-LMs).}
\textsc{Eso-LMs} fuse AR and MDM paradigms using causal attention, enabling
exact likelihood computation and KV caching for MDMs~\cite{sahoo2026esoteric}.
They achieve low
perplexities among diffusion models on One Billion Words~\cite[LM1B;][]{chelba-etal-2014-OneBillionWord} and OpenWebText~\cite[OWT;][]{gokaslan-etal-2019-OpenwebtextCorpus} datasets.

\paragraph{ReFusion.}
\textsc{ReFusion} introduces slot-level parallel decoding, elevating
generation from tokens to fixed-length slots~\cite{li2026refusion}.
It interleaves inter-slot diffusion-based selection with intra-slot
autoregressive infilling, reordering newly generated slots ahead of
remaining masks after each iteration.
This design unlocks full KV
cache reuse and reduces learning complexity from an intractable token
combination space to a manageable slot-level permutation space.
However, \textsc{ReFusion} inference is tied only for prefix completion tasks.

\section{Zarya Architecture}

\subsection{Model Backbone and Configuration}

\textsc{Zarya} wraps a Qwen3 backbone~\cite{qwen3technicalreport} with a
custom \texttt{Zarya} class registered with Hugging Face's
\texttt{AutoModel} and \texttt{AutoConfig} systems.
We initialize \textsc{Zarya} from the Qwen3-0.6B, Qwen3-1.7B, and Qwen3-4B checkpoints, respectively, and fine-tune it for 1 epoch on a diverse 32M-sample instructional SFT dataset (approximately 37.7B tokens) covering
mathematics, coding, and general instruction-following tasks.
We release \textsc{Zarya} in three sizes (see \autoref{tab:architecture}).

Because the Qwen3 backbone retains causal attention, masked positions cannot attend to future masked positions.
Thus, the diffusion objective used by \textsc{Zarya} is a causal masked-reconstruction objective rather than the fully bidirectional masked-token objective commonly used in masked diffusion language models.
The reordering of visible and masked positions ensures that all masked positions can attend to the visible prefix while preserving the causal attention pattern and KV-cache compatibility.

\begin{table}[t]
  \caption{Model architecture of Zarya models\label{tab:architecture}}
  \small
  \centering
  \begin{tabularx}{\linewidth}{%
      @{}l%
      >{\centering\arraybackslash}X%
      >{\centering\arraybackslash}X%
      >{\centering\arraybackslash}X%
      >{\centering\arraybackslash}X%
      >{\centering\arraybackslash}X%
      >{\centering\arraybackslash}X@{}%
    }
    \toprule
    \textbf{Model}  & \textbf{Layers} & \textbf{Heads (\mbox{Q / KV})} & \textbf{Tie Embedding} & \textbf{Hidden Size} & \textbf{Intermediate Size} & \textbf{Context Length}  \\
    \midrule
    Zarya-0.6B  & 28 & 16 / 8 & Yes & 1024 & 3072 & 2K   \\
    Zarya-1.7B  & 28 & 16 / 8 & Yes & 2048 & 6144 & 2K   \\
    Zarya-4B  & 36 & 32 / 8 & Yes & 2560 & 9728 & 2K  \\
    \bottomrule
  \end{tabularx}
\end{table}

The architecture
is parameterized with diffusion-specific hyperparameters, which extends common AR model parameters:

\begin{itemize}
\item $\alpha_0$ (default 0.25) and $\epsilon$ (default 0.001):
  parameters of the linear noise schedule $\alpha_t = \alpha_0 (1 - t)$;
\item $\lambda = \texttt{diffusion\_loss\_proportion}$ (default 0.5):
  weighting between MDM and AR losses;
\item \texttt{sequential\_shuffle} / \texttt{diffusion\_shuffle}:
  control slot/token shuffling for each phase;
\item \texttt{ordered\_sampling}: monotonically increases $p_\text{mask}$
  left-to-right;
\item \texttt{grouped\_noise} and \texttt{max\_span\_length}: enable
  Prefix Completion, Fill-in-the-Prefix, Fill-in-the-Middle, and random-span
  masking patterns;
\item \texttt{noise\_sorting}: reorders tokens by mask/unmask state
  before the forward pass.
\end{itemize}

\section{Training}

\subsection{Slotted Training}
\label{sec:slotted-training}

When \texttt{slotted\_training=True}, the \texttt{forward\_process()}
transforms each batch as follows (this regime is heavily inspired by~\citet{li2026refusion}):
\begin{enumerate}
\item \textbf{Slot partitioning:} Each answer is split into slots of
  size $\texttt{slot\_size}$ (from $\texttt{slot\_size\_set}$, e.g., for released model checkpoints
  $[2, 4, 8, 16, 32, 64]$).
\item \textbf{Mask sampling:} For each example in batch, sample mask probability (\(p_\text{mask}\)) uniformly.
  This determines what fraction of slots  will be masked (treated as the diffusion task).
\item \textbf{Slot assignment:}
  \begin{itemize}
  \item \emph{AR slots}: tokens
    remain visible; the model predicts the next token within each
    slot, yielding $\mathcal{L}_\text{seq}$.
  \item \emph{MDM slots}: all tokens
    are replaced with the special $\langle\texttt{mdm\_mask}\rangle$
    token; the model reconstructs the original tokens, yielding
    $\mathcal{L}_\text{dif}$.
    Each token inside slot assigned $p_\text{mask}$ value to be used in per-token normalization as $\frac{1}{p_\text{mask}}$ weighting.
  \end{itemize}
\end{enumerate}

The final loss is a linear combination:

\begin{equation}
\mathcal{L} = \lambda \cdot \mathcal{L}_\text{dif} + (1 - \lambda)
\cdot \mathcal{L}_\text{seq},
\end{equation}
where $\lambda$ is diffusion loss proportion.

This design enables the model to simultaneously learn next-token
prediction (AR) and masked-token reconstruction (MDM) on the
\emph{same} input, with explicit slot boundaries providing a
structured inductive bias.

Optional flags include:
\begin{itemize}
  \item \texttt{ordered\_sampling}: inside each slot modify per-token $p_\text{mask}$ so that it increases left-to-right, as $\mathcal{L}_\text{dif}$ is scaled with the value of $\frac{1}{p_\text{mask}}$, this makes tne model learn that correct prediction of the slot beginning tokens is more important;
\end{itemize}

\subsubsection{Slot-Size Curriculum}

\begin{figure}[t]
  \centering
  \includegraphics[width=0.5\textwidth]{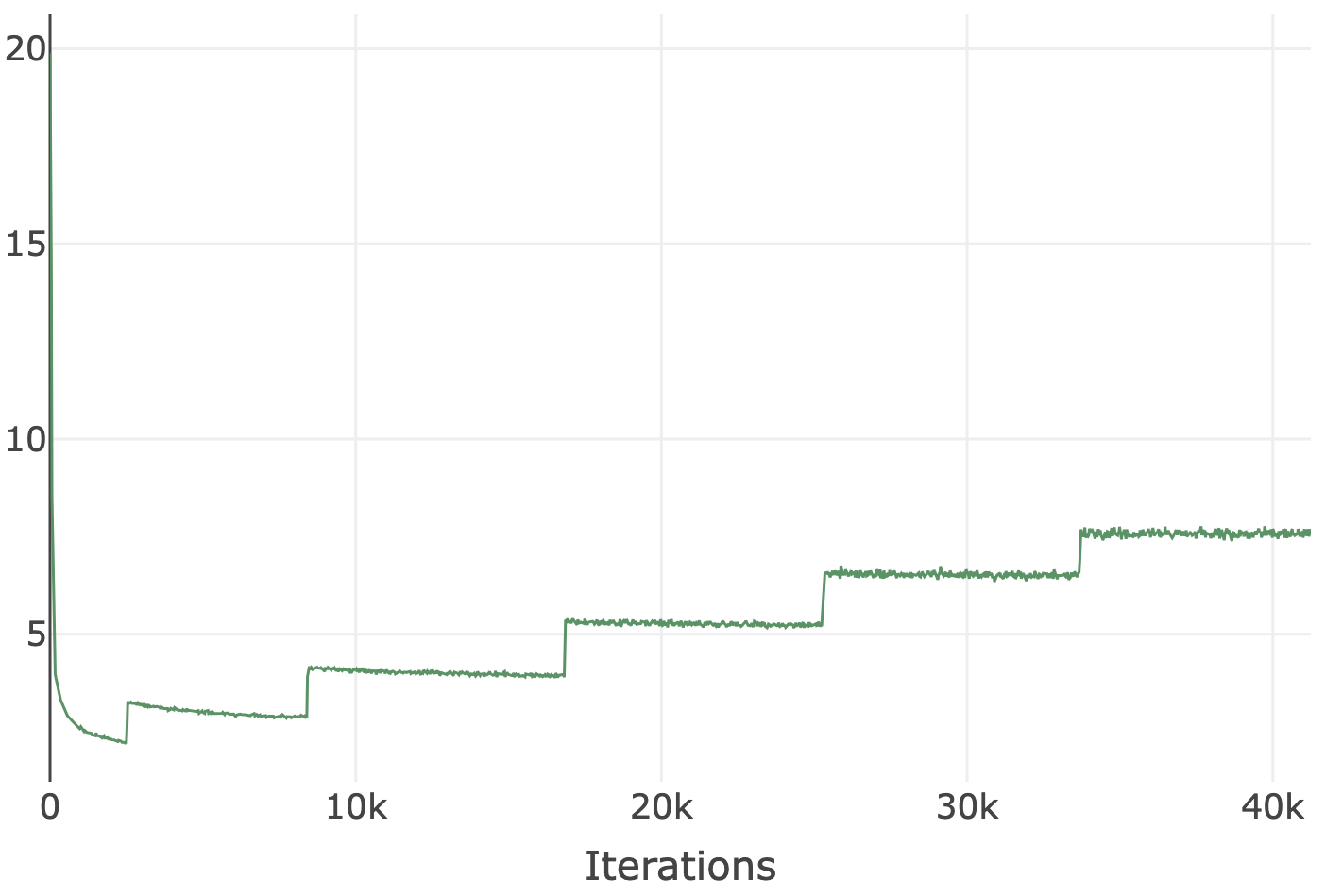}
  \caption{Loss curve during training of Zarya-4B model. Step-by-step increasing of slot size causes increase of loss too}
  \label{fig:loss-train-4b}
\end{figure}

Slot sizes evolve during training via \texttt{slot\_step\_borders}
(epoch or step thresholds).
For example, with
\begin{align*}
\texttt{slot\_size\_set} &= [2, 4, 8, 16, 32, 64], \\
\texttt{slot\_step\_borders} &= [0.06, 0.2, 0.4, 0.6, 0.8, 1.0],
\end{align*}
the slot size gradually increases from 2 to 64 over the course of
training.
This curriculum eases the model from fine-grained AR
learning (small slots, many predictions) to coarse-grained diffusion
learning (large slots, holistic reconstruction), providing a smooth
transition between paradigms.
Worth noting that increase in slot size also increases calculated loss (see \autoref{fig:loss-train-4b}).

\subsection{Non-Slotted Training}

When \texttt{slotted\_training=False}, the model runs
objective over raw sequences without slot partitioning (this regime is heavily inspired by~\citet{sahoo2026esoteric}):
\begin{enumerate}
\item For each example in batch, \textbf{sample noise level} (\(t\)) uniformly.
\item \textbf{Construct noisy input}. Noise schedule is $\alpha_t = \alpha_0 (1 - t)$ and the per-token mask probability is $p_\text{mask} = 1 - \alpha_t$.
\item Run \textbf{two separate forward passes}:
  \begin{itemize}
  \item \emph{sequential phase}: clean sequence $x_0$, predict only tokens at
    masked positions (others set to \texttt{ignore\_index}).
  \item \emph{diffusion phase}: noisy sequence $x_t$, reconstruct
    masked tokens.
  \end{itemize}
\item Final loss is linear combination of losses at sequential and diffusion phases, calculated the same way as in \autoref{sec:slotted-training}.
\end{enumerate}

Optional flags include:
\begin{itemize}
\item \texttt{noise\_sorting}: reorders tokens by mask/unmask state
  before both forward passes, with logits permuted back afterward;
\item \texttt{grouped\_noise}: masks contiguous spans (Prefix Completion, Fill-in-the-Prefix, Fill-in-the-Middle, random spans) up to
  \texttt{max\_span\_length};
\item \texttt{ordered\_sampling}: offsets $t$ per token position so
  that $p_\text{mask}$ increases left-to-right.
\end{itemize}

\section{Inference: Dual-Mode Unified Interface}

\textsc{Zarya} provides two distinct decoding paradigms through a
single \texttt{model.generate()} call, routed via
\texttt{generation\_config.slotted\_generation}.
Critically, the
training and inference regimes are fully decoupled: a model trained
with \texttt{slotted\_training=False} can still be deployed with
\texttt{slotted\_generation=True}, and vice versa.

\subsection{Mode A: MDM Sampling}
\label{sec:mdm-sampling}

When \texttt{slotted\_generation=False}, inference code
executes the first-hitting denoising process~\cite{zheng-etal-2025-MaskedDiffusionModels, sahoo2026esoteric}.
Starting from
the prompt padded with $\langle\texttt{mdm\_mask}\rangle$ tokens up
to \texttt{max\_length}, the model iteratively reveals tokens:
\begin{enumerate}
\item \textbf{Mask budget planning:}
  \texttt{\_tokens\_unmasked\_per\_step()} determines how many masks
  to reveal per step. $\alpha_0$ is the expected fraction of masked tokens generated with diffusion process.
  \begin{itemize}
  \item If number of discretization steps is set to $T > 0$: exactly $T$ diffusion steps are used. Binomial distribution is used to calculate number of masked tokens to denoise through diffusion process and tokens thats left after that to denoise
  sequentially (this mode is heavily inspired by~\citet[Appendix B.5]{sahoo2026esoteric}).
  \item If number of discretization steps is set to $T = 0$ (ignoring noise calculations): $T$ steps are auto-calculated as $\frac{1}{4}$ of masked tokens.
  \end{itemize}
\item \textbf{Reordering:} Input sequence is reordered so that masked tokens are always after unmasked.
\item \textbf{Per-step sampling from categorical distribution:} At each step, the model receives
  the progressively filled sequence (with KV cache reuse) and yields
  logits for masked positions.
  Gumbel noise~\parencites{gumbel1935valeurs}[Appendix F]{zheng-etal-2025-MaskedDiffusionModels} is added for categorical
  sampling, and standard sampling parameters (\texttt{temperature},
  \texttt{top\_p}, \texttt{repetition\_penalty}) are honored.
\item \textbf{Restoration:} After all steps, the sequence is restored to the original token order.
\end{enumerate}

The KV cache can be reused because the sequence is reordered so that tokens whose values are fixed at a given denoising step precede the remaining masked positions.
Under causal attention, the cached prefix states therefore remain unchanged when masked positions are progressively filled.
This property would not hold for a bidirectional masked-diffusion attention pattern, where changing any previously masked token could affect the representations of other masked positions.

\subsection{Mode B: Slotted Speculative Decoding}
\label{sec:slotted-speculative}

When \texttt{slotted\_generation=True}, inference code
achieves parallelization by elevating decoding units from tokens to
slots, fully reusing KV cache to avoid recomputation.
This mode follows the \textsc{ReFusion} paradigm~\cite{li2026refusion}:
\begin{enumerate}
 \item \textbf{Reordering:} Before the first forward pass, input sequence is reordered so that masked tokens are always after unmasked.
\item \textbf{Block construction:} \texttt{max\_new\_tokens} count of masked tokens is
  divided into \texttt{serial\_num\_blocks} blocks of length
  $\texttt{block\_size} = \lfloor \texttt{max\_new\_tokens} /
  \texttt{serial\_num\_blocks} \rfloor$.
  Within each block, tokens
  are grouped into slots of size $\texttt{slot\_size}$.
  If
  \texttt{max\_new\_tokens} is small,
  \texttt{serial\_num\_blocks} is forced to 1 to prevent zero-length
  blocks.
\item \textbf{Draft phase:} An MDM forward pass drafts tokens for all
  slots in the current block in parallel.
  The confidence of each slot
  is estimated as the probability of its first token.
\item \textbf{Sampling from categorical distribution:} If \texttt{temperature} is positive, apply Gumbel noise to logits for sampling from categorical distributions.
\item \textbf{Slot selection:} Slots with confidence exceeding
  \texttt{slot\_threshold} are accepted immediately.
  If no slots are confident enough, select the most confident one, so we always have at least one slot to process further.
\item \textbf{Verification phase:} Selected slots undergo an AR
  verification forward pass.
  Tokens with probability exceeding
  \texttt{token\_threshold} are accepted; those below are iteratively
  refined in a speculative loop.
\item \textbf{KV cache update:} The cache is updated incrementally
  with accepted tokens, avoiding recomputation for verified
  prefixes.
\item \textbf{Restoration:} After all steps, the sequence is restored to the original token order.
\end{enumerate}

\subsection{Parameter Decoupling Summary}

\autoref{tab:param-scope} summarizes the scope of key configuration
parameters.

\begin{table}[t]
\centering
\caption{Scope of configuration parameters across training and inference modes.}
\label{tab:param-scope}
\begin{tabular}{lcc}
\toprule
\textbf{Parameter} & \textbf{Training} & \textbf{Inference} \\
\midrule
\texttt{slotted\_training} & \checkmark & --- \\
\texttt{slot\_size\_set} / \texttt{slot\_step\_borders} & \checkmark & --- \\
\texttt{diffusion\_loss\_proportion} ($\lambda$) & \checkmark & --- \\
\texttt{noise\_sorting} & \checkmark & --- \\
\texttt{ordered\_sampling} & \checkmark & --- \\
\texttt{grouped\_noise} / \texttt{max\_span\_length} & \checkmark & --- \\
\texttt{add\_loss\_path} / \texttt{scale\_by\_batch} & \checkmark & --- \\
\midrule
\texttt{slotted\_generation} & --- & \checkmark (routes Mode A/B) \\
$T$ (diffusion steps) & --- & \checkmark (Mode A) \\
\texttt{sequential\_shuffle} / \texttt{diffusion\_shuffle} & \checkmark & \checkmark (Mode A only) \\
\texttt{slot\_size} / \texttt{serial\_num\_blocks} & --- & \checkmark (Mode B) \\
\texttt{slot\_threshold} / \texttt{token\_threshold} & --- & \checkmark (Mode B) \\
\texttt{temperature} / \texttt{top\_p} / \texttt{repetition\_penalty} & --- & \checkmark (both modes) \\
\bottomrule
\end{tabular}
\end{table}

\section{Evaluation}

We evaluate on standard benchmarks including GSM8K (mathematical
reasoning)~\cite{cobbe2021trainingverifierssolvemath}, HellaSwag (commonsense reasoning)~\cite{zellers-etal-2019-hellaswag}, IFEval (instruction
following)~\cite{zhou-etal-2023-InstructionFollowingEvaluation}, and MBPP (code generation)~\cite{austin-etal-2021-ProgramSynthesisLarge}.
Preliminary results for
Zarya-0.6B are shown in \autoref{tab:results}.
Hardware info:
GPU A100;
GPU driver CUDA version 13.2;
GPU driver version 595.71.05;
Docker info:
Torch: 2.9.0+cu128; Transformers: 5.12.1; CUDNN in torch: 91002; lm-eval 0.4.12;
Inference info: BF16, apply chat template, \verb|slotted_generation=true|, \verb|slot_size=16|,
\verb|serial_num_blocks=4|, \verb|slot_threshold=0.9|, \verb|token_threshold=0.4|

\begin{table}[tb]
  \centering
  \caption{Preliminary evaluation results for Zarya-0.6B}
  \label{tab:results}
  \small
  \begin{tabular}{lllllllll}
    \toprule
    \textbf{Tasks} & \textbf{Version} & \textbf{Filter} & \textbf{n-shot} & \textbf{Metric} & \textbf{} & \textbf{Value} & \textbf{} & \textbf{Stderr} \\
    \midrule
    gsm8k & 3 & flexible-extract & 5 & exact\_match & ↑ & 0.2646 & ± & 0.0122 \\
    ~ & ~ & strict-match & 5 & exact\_match & ↑ & 0.2646 & ± & 0.0122 \\
    hellaswag & 1 & none & 0 & acc & ↑ & 0.3526 & ± & 0.0048 \\
    ~ & ~ & none & 0 & acc\_norm & ↑ & 0.4270 & ± & 0.0049 \\
    ifeval & 4 & none & 0 & inst\_level\_loose\_acc & ↑ & 0.5372 & ± & N/A \\
    ~ & ~ & none & 0 & inst\_level\_strict\_acc & ↑ & 0.5108 & ± & N/A \\
    ~ & ~ & none & 0 & prompt\_level\_loose\_acc & ↑ & 0.4177 & ± & 0.0212 \\
    ~ & ~ & none & 0 & prompt\_level\_strict\_acc & ↑ & 0.3993 & ± & 0.0211 \\
    mbpp & 1 & none & 3 & pass\_at\_1 & ↑ & 0.1500 & ± & 0.0160 \\
    mbpp\_plus & 1 & none & 3 & pass\_at\_1 & ↑ & 0.2249 & ± & 0.0215 \\
    \bottomrule
  \end{tabular}
\end{table}

\section{Conclusion}

We introduced \textsc{Zarya}, a hybrid AR--MDM language model that
combines slotted training, dual-mode
inference, and extensive configurability.
By structuring data into
slots and gradually increasing slot size during training,
\textsc{Zarya} achieves a principled integration of autoregressive
and diffusion paradigms.
The unified inference interface
supports both MDM sampling and slotted speculative
decoding mode with full KV cache reuse.
\textsc{Zarya} represents a
step toward flexible, efficient language models that leverage the
strengths of both generation paradigms.
We release models at three
sizes and provide a comprehensive training framework
for further research.

\section*{Limitations}

While \textsc{Zarya} demonstrates strong potential, several
limitations remain:

\begin{enumerate}
\item \textbf{Threshold sensitivity:} The performance of slotted
  speculative decoding depends critically on slot selection threshold
  and token selection threshold.
  Poorly tuned thresholds can lead to
  low acceptance rates or quality degradation.
  An
  adaptive thresholding mechanism is needed.
\item \textbf{Mode-specific ignorance:} Researchers must remember
  that \texttt{ordered\_sampling} and \texttt{noise\_sorting} configuration options do
  \emph{nothing} during inference.
  If these are tuned heavily during
  training, users may see no effect at inference time.
\end{enumerate}

\section*{Acknowledgments}

We thank the open-source community for providing the foundation upon
which this work is built.

\phantomsection
\addcontentsline{toc}{section}{References}
\printbibliography

@InProceedings{arriola-etal-2025-BlockDiffusionInterpolating,
  author    = {Arriola, Marianne and Sahoo, Subham Sekhar and Gokaslan, Aaron and Yang, Zhihan and Qi, Zhixuan and Han, Jiaqi and Chiu, Justin T. and Kuleshov, Volodymyr},
  booktitle = {The Thirteenth International Conference on Learning Representations},
  title     = {Block Diffusion: Interpolating Between Autoregressive and Diffusion Language Models},
  url       = {https://openreview.net/forum?id=tyEyYT267x},
  year      = {2025},
}

@InProceedings{brown2020language,
  author    = {Brown, Tom and Mann, Benjamin and Ryder, Nick and Subbiah, Melanie and Kaplan, Jared D. and Dhariwal, Prafulla and Neelakantan, Arvind and Shyam, Pranav and Sastry, Girish and Askell, Amanda and Agarwal, Sandhini and Herbert-Voss, Ariel and Krueger, Gretchen and Henighan, Tom and Child, Rewon and Ramesh, Aditya and Ziegler, Daniel and Wu, Jeffrey and Winter, Clemens and Hesse, Chris and Chen, Mark and Sigler, Eric and Litwin, Mateusz and Gray, Scott and Chess, Benjamin and Clark, Jack and Berner, Christopher and McCandlish, Sam and Radford, Alec and Sutskever, Ilya and Amodei, Dario},
  booktitle = {Advances in Neural Information Processing Systems},
  title     = {Language Models are Few-Shot Learners},
  editor    = {Larochelle, H. and Ranzato, M. and Hadsell, R. and Balcan, M. F. and Lin, H.},
  pages     = {1877--1901},
  publisher = {Curran Associates, Inc.},
  url       = {https://proceedings.neurips.cc/paper_files/paper/2020/file/1457c0d6bfcb4967418bfb8ac142f64a-Paper.pdf},
  volume    = {33},
  year      = {2020},
}

@InProceedings{cai2024medusa,
  author    = {Tianle Cai and
Yuhong Li and
Zhengyang Geng and
Hongwu Peng and
Jason D. Lee and
Deming Chen and
Tri Dao},
  booktitle = {Forty-first International Conference on Machine Learning, {ICML} 2024,
Vienna, Austria, July 21-27, 2024},
  title     = {Medusa: Simple {LLM} Inference Acceleration Framework with Multiple
Decoding Heads},
  publisher = {OpenReview.net},
  url       = {https://openreview.net/forum?id=PEpbUobfJv},
  year      = {2024},
}

@Misc{chen2023speculative,
  author        = {Chen, Charlie and Borgeaud, Sebastian and Irving, Geoffrey and Lespiau, Jean-Baptiste and Sifre, Laurent and Jumper, John},
  title         = {Accelerating Large Language Model Decoding with Speculative Sampling},
  eprint        = {2302.01318},
  url           = {https://arxiv.org/abs/2302.01318},
  archiveprefix = {arXiv},
  primaryclass  = {cs.CL},
  year          = {2023},
}

@InProceedings{li2026refusion,
  author    = {Li, Jia-Nan and Guan, Jian and Wu, Wei and Li, Chongxuan},
  booktitle = {International Conference on Learning Representations},
  title     = {{ReFusion}: A Diffusion Large Language Model with Parallel Autoregressive Decoding},
  editor    = {Vondrick, C. and Hariharan, B. and Raffel, C. and Pinto, L. and Yang, D. and Faust, A.},
  pages     = {53846--53869},
  url       = {https://proceedings.iclr.cc/paper_files/paper/2026/file/585979c057a1b30796cf317063559638-Paper-Conference.pdf},
  volume    = {2026},
  year      = {2026},
}

@Misc{li-etal-2026-SurveyDiffusionLanguage,
  author        = {Li, Tianyi and Chen, Mingda and Guo, Bowei and Shen, Zhiqiang},
  title         = {A Survey on Diffusion Language Models},
  eprint        = {2508.10875},
  url           = {https://arxiv.org/abs/2508.10875},
  archiveprefix = {arXiv},
  primaryclass  = {cs.CL},
  year          = {2026},
}

@Misc{openai2023gpt4,
  author        = {OpenAI},
  title         = {GPT-4 Technical Report},
  eprint        = {2303.08774},
  url           = {https://arxiv.org/abs/2303.08774},
  archiveprefix = {arXiv},
  primaryclass  = {cs.CL},
  year          = {2023},
}

@InProceedings{sahoo2026esoteric,
  author    = {Sahoo, Subham Sekhar and Yang, Zhihan and Akhauri, Yash and Liu, Johnna and Singh, Deepansha and Cheng, Zhoujun and Liu, Zhengzhong and Xing, Eric P. and Thickstun, John and Vahdat, Arash},
  booktitle = {ICLR 2026 Workshop on Multimodal Intelligence},
  title     = {Esoteric Language Models: Bridging Autoregressive and Masked Diffusion LLMs},
  url       = {https://openreview.net/forum?id=CKrPJveQIr},
  year      = {2026},
}

@Misc{touvron2023llama2,
  author        = {Touvron, Hugo and Martin, Louis and Stone, Kevin and Albert, Peter and Almahairi, Amjad and Babaei, Yasmine and Bashlykov, Nikolay and Batra, Soumya and Bhargava, Prajjwal and Bhosale, Shruti and Bikel, Dan and Blecher, Lukas and Ferrer, Cristian Canton and Chen, Moya and Cucurull, Guillem and Esiobu, David and Fernandes, Jude and Fu, Jeremy and Fu, Wenyin and Fuller, Brian and Gao, Cynthia and Goswami, Vedanuj and Goyal, Naman and Hartshorn, Anthony and Hosseini, Saghar and Hou, Rui and Inan, Hakan and Kardas, Marcin and Kerkez, Viktor and Khabsa, Madian and Kloumann, Isabel and Korenev, Artem and Koura, Punit Singh and Lachaux, Marie-Anne and Lavril, Thibaut and Lee, Jenya and Liskovich, Diana and Lu, Yinghai and Mao, Yuning and Martinet, Xavier and Mihaylov, Todor and Mishra, Pushkar and Molybog, Igor and Nie, Yixin and Poulton, Andrew and Reizenstein, Jeremy and Rungta, Rashi and Saladi, Kalyan and Schelten, Alan and Silva, Ruan and Smith, Eric Michael and Subramanian, Ranjan and Tan, Xiaoqing Ellen and Tang, Binh and Taylor, Ross and Williams, Adina and Kuan, Jian Xiang and Xu, Puxin and Yan, Zheng and Zarov, Iliyan and Zhang, Yuchen and Fan, Angela and Kambadur, Melanie and Narang, Sharan and Rodriguez, Aurelien and Stojnic, Robert and Edunov, Sergey and Scialom, Thomas},
  title         = {Llama 2: Open Foundation and Fine-Tuned Chat Models},
  eprint        = {2307.09288},
  url           = {https://arxiv.org/abs/2307.09288},
  archiveprefix = {arXiv},
  primaryclass  = {cs.CL},
  year          = {2023},
}

@Misc{qwen3technicalreport,
  author        = {Yang, An and Li, Anfeng and Yang, Baosong and Zhang, Beichen and Hui, Binyuan and Zheng, Bo and Yu, Bowen and Gao, Chang and Huang, Chengen and Lv, Chenxu and Zheng, Chujie and Liu, Dayiheng and Zhou, Fan and Huang, Fei and Hu, Feng and Ge, Hao and Wei, Haoran and Lin, Huan and Tang, Jialong and Yang, Jian and Tu, Jianhong and Zhang, Jianwei and Yang, Jianxin and Yang, Jiaxi and Zhou, Jing and Zhou, Jingren and Lin, Junyang and Dang, Kai and Bao, Keqin and Yang, Kexin and Yu, Le and Deng, Lianghao and Li, Mei and Xue, Mingfeng and Li, Mingze and Zhang, Pei and Wang, Peng and Zhu, Qin and Men, Rui and Gao, Ruize and Liu, Shixuan and Luo, Shuang and Li, Tianhao and Tang, Tianyi and Yin, Wenbiao and Ren, Xingzhang and Wang, Xinyu and Zhang, Xinyu and Ren, Xuancheng and Fan, Yang and Su, Yang and Zhang, Yichang and Zhang, Yinger and Wan, Yu and Liu, Yuqiong and Wang, Zekun and Cui, Zeyu and Zhang, Zhenru and Zhou, Zhipeng and Qiu, Zihan},
  title         = {Qwen3 Technical Report},
  eprint        = {2505.09388},
  url           = {https://arxiv.org/abs/2505.09388},
  archiveprefix = {arXiv},
  primaryclass  = {cs.CL},
  year          = {2025},
}

@Misc{cobbe2021trainingverifierssolvemath,
  author        = {Cobbe, Karl and Kosaraju, Vineet and Bavarian, Mohammad and Chen, Mark and Jun, Heewoo and Kaiser, Lukasz and Plappert, Matthias and Tworek, Jerry and Hilton, Jacob and Nakano, Reiichiro and Hesse, Christopher and Schulman, John},
  title         = {Training Verifiers to Solve Math Word Problems},
  eprint        = {2110.14168},
  url           = {https://arxiv.org/abs/2110.14168},
  archiveprefix = {arXiv},
  primaryclass  = {cs.LG},
  year          = {2021},
}

@InProceedings{zellers-etal-2019-hellaswag,
  author    = {Zellers, Rowan and Holtzman, Ari and Bisk, Yonatan and Farhadi, Ali and Choi, Yejin},
  booktitle = {Proceedings of the 57th Annual Meeting of the Association for Computational Linguistics},
  title     = {{H}ella{S}wag: Can a Machine Really Finish Your Sentence?},
  doi       = {10.18653/v1/P19-1472},
  editor    = {Korhonen, Anna and Traum, David and M{\`a}rquez, Llu{\'i}s},
  pages     = {4791--4800},
  publisher = {Association for Computational Linguistics},
  url       = {https://aclanthology.org/P19-1472/},
  address   = {Florence, Italy},
  month     = jul,
  year      = {2019},
}

@Misc{austin-etal-2021-ProgramSynthesisLarge,
  author        = {Austin, Jacob and Odena, Augustus and Nye, Maxwell and Bosma, Maarten and Michalewski, Henryk and Dohan, David and Jiang, Ellen and Cai, Carrie and Terry, Michael and Le, Quoc and Sutton, Charles},
  title         = {Program Synthesis with Large Language Models},
  eprint        = {2108.07732},
  url           = {https://arxiv.org/abs/2108.07732},
  archiveprefix = {arXiv},
  primaryclass  = {cs.PL},
  year          = {2021},
}

@Misc{chelba-etal-2014-OneBillionWord,
  author        = {Chelba, Ciprian and Mikolov, Tomas and Schuster, Mike and Ge, Qi and Brants, Thorsten and Koehn, Phillipp and Robinson, Tony},
  title         = {One Billion Word Benchmark for Measuring Progress in Statistical Language Modeling},
  eprint        = {1312.3005},
  url           = {https://arxiv.org/abs/1312.3005},
  archiveprefix = {arXiv},
  primaryclass  = {cs.CL},
  year          = {2014},
}

@Misc{gokaslan-etal-2019-OpenwebtextCorpus,
  author = {Gokaslan, Aaron and Cohen, Vanya and Pavlick, Ellie and Tellex, Stefanie},
  title  = {{OpenWebText} Corpus},
  url    = {http://Skylion007.github.io/OpenWebTextCorpus},
  year   = {2019},
}

@InProceedings{zheng-etal-2025-MaskedDiffusionModels,
  author    = {Zheng, Kaiwen and Chen, Yongxin and Mao, Hanzi and Liu, Ming-Yu and Zhu, Jun and Zhang, Qinsheng},
  booktitle = {International Conference on Learning Representations},
  title     = {Masked Diffusion Models are Secretly Time-Agnostic Masked Models and Exploit Inaccurate Categorical Sampling},
  editor    = {Yue, Y. and Garg, A. and Peng, N. and Sha, F. and Yu, R.},
  pages     = {63186--63227},
  url       = {https://proceedings.iclr.cc/paper_files/paper/2025/file/9e3b203e72c4e058de26d02a92a81844-Paper-Conference.pdf},
  volume    = {2025},
  year      = {2025},
}

@Article{gumbel1935valeurs,
  author       = {Gumbel, Emil Julius},
  title        = {Les valeurs extr{\^e}mes des distributions statistiques},
  number       = {2},
  origlanguage = {fr},
  pages        = {115--158},
  url          = {https://www.numdam.org/item/AIHP_1935__5_2_115_0/},
  volume       = {5},
  booktitle    = {Annales de l'institut Henri Poincar{\'e}},
  publisher    = {INSTITUT HENRI POINCAR\'E ET LES PRESSES UNIVERSITAIRES DE FRANCE},
  year         = {1935},
}

@Misc{zhou-etal-2023-InstructionFollowingEvaluation,
  author        = {Zhou, Jeffrey and Lu, Tianjian and Mishra, Swaroop and Brahma, Siddhartha and Basu, Sujoy and Luan, Yi and Zhou, Denny and Hou, Le},
  title         = {Instruction-Following Evaluation for Large Language Models},
  eprint        = {2311.07911},
  url           = {https://arxiv.org/abs/2311.07911},
  archiveprefix = {arXiv},
  primaryclass  = {cs.CL},
  year          = {2023},
}

\end{document}